\documentclass[10pt,twocolumn,letterpaper]{article}

\usepackage[pagenumbers]{cvpr} 
\usepackage{tcolorbox}
\usepackage{lipsum} 
\usepackage[utf8]{inputenc}
\usepackage{graphicx}
\usepackage{amsmath}
\usepackage{amssymb}
\usepackage{booktabs}
\usepackage[table]{xcolor}
\usepackage{tabularx} 
\usepackage{threeparttable}

\tcbuselibrary{skins}
\usepackage{multirow}
\usepackage{makecell}
\usepackage[dvipsnames]{xcolor}
\usepackage{url}
\usepackage[toc,page]{appendix}

\usepackage{tcolorbox}
\tcbuselibrary{skins}

\definecolor{cvprblue}{rgb}{0.21,0.49,0.74}
\usepackage[pagebackref,breaklinks,colorlinks,allcolors=cvprblue]{hyperref}

\def\paperID{} 
\def\confName{CVPR}
\def\confYear{2026}

\title{Nano\mbox{-}EmoX: 

Unifying Multimodal Emotional Intelligence from Perception to Empathy}

\author{
  Jiahao Huang$^{1}$ \quad Fengyan Lin$^{1}$ \quad Xuechao Yang$^{2}$ \quad Chen Feng$^{4}$ \quad Kexin Zhu$^{4}$\\
  Xu Yang$^{3*}$ \quad Zhide Chen$^{1*}$\\
  $^1$Fujian Normal University \quad $^2$RMIT University \quad $^3$Minjiang University \quad $^4$Independent Researcher\\
}

\begin{document}
\maketitle
\renewcommand{\thefootnote}{\fnsymbol{footnote}}
\footnotetext[1]{Corresponding authors}
\begin{abstract}
The development of affective multimodal language models (MLMs) has long been constrained by a gap between low-level perception and high-level interaction, leading to fragmented affective capabilities and limited generalization. To bridge this gap, we propose a cognitively inspired three-level hierarchy that organizes affective tasks according to their cognitive depth—perception, understanding, and interaction—and provides a unified conceptual foundation for advancing affective modeling. Guided by this hierarchy, we introduce \textbf{Nano\mbox{-}EmoX}, a small-scale multitask MLM, and \textbf{P2E} (\textbf{P}erception-\textbf{to}-\textbf{E}mpathy), a curriculum-based training framework. Nano\mbox{-}EmoX integrates a suite of omni-modal encoders, including an enhanced facial encoder and a fusion encoder, to capture key multimodal affective cues and improve cross-task transferability. The outputs are projected into a unified language space via heterogeneous adapters, empowering a lightweight language model to tackle diverse affective tasks. Concurrently, P2E progressively cultivates emotional intelligence by aligning rapid perception with chain-of-thought-driven empathy.
To the best of our knowledge, Nano\mbox{-}EmoX is the first compact MLM (2.2B) to unify six core affective tasks across all three hierarchy levels, achieving state-of-the-art or highly competitive performance across multiple benchmarks, demonstrating excellent efficiency and generalization. The code is available at \url{https://github.com/waHAHJIAHAO/Nano-EmoX}.

\end{abstract}    
\section{Introduction}
\label{sec:intro}

\begin{figure}[t]
    \centering
    \includegraphics[width=1.0\linewidth]{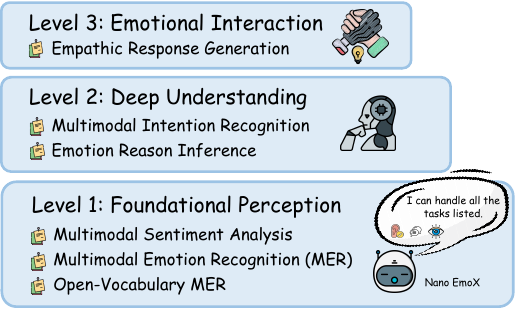}
    \caption{This framework organizes tasks by increasing cognitive depth: (1) \emph{Perception} for direct recognition of emotional cues; (2) \emph{Understanding} for inferring emotional causality and context; and (3) \emph{Emotional Interaction} for establishing an emotional connection with humans. Please refer to the appendix for details.}
    \label{fig:task_hierarchy}
\end{figure}

To advance human-centric AI, systems must move beyond simple emotion perception toward holistic emotional intelligence, a unified continuum from perception to interaction~\cite{JIANG2024100078,picard2000affective}. However, the current landscape of affective computing remains a vast yet fragmented collection of tasks, lacking a coherent structure to guide systematic progress or to assess a model's true emotional maturity.
Motivated by the Perception–Action Model \cite{preston2002empathy}, we introduce a three-level cognitive hierarchy for organizing affective tasks. As illustrated in \cref{fig:task_hierarchy}, the hierarchy arranges emotional tasks by cognitive depth, ascending from foundational perception to deeper understanding and emotional interaction, mapping each task to a progressively more advanced level of affective processing.
%
%
%
%
%
%

Viewing the field through this hierarchical lens clarifies its historical trajectory. Early research—from unimodal~\cite{sabour2022cem,plaza-del-arco-2021-multi,cai2021speech,li2019attentive,hallmen2024unimodal,savchenko2021facial} to multimodal pipelines~\cite{peng2024carat,akhtar2019multi,10205489,11092537,10204378,10203083,10203999}—primarily addressed challenges at a single level. The advent of Large Language Models (LLMs)~\cite{fei-etal-2023-reasoning,wu2023improving,liu2024emollms} and Multimodal Language Models (MLMs)~\cite{yang2024emollm,xie2024emovit,huang2025emotion,cheng2024emotion,lian2025affectgpt,yang2025omni,zhao2025r1} catalyzed a significant shift, enabling models to master analytical tasks at the understanding level. More recently, this progress has culminated in pioneering efforts toward the interaction stratum~\cite{zhang2025towards,liu2024speak}.

\begin{table}[t]
\centering
\caption{Comparison with representative LLM-based methods under different task settings, our method unifies six core affective tasks with a smaller parameter scale.}
\label{tab:cap_compare}
\resizebox{\linewidth}{!}{ 
\renewcommand{\arraystretch}{1.3} 
\begin{tabular}{>{\large}c>{\large}c>{\large}cccccccc}   
\toprule
\textbf{Models} & \textbf{Scale} & \textbf{Hierarchy} & 
\rotatebox[origin=c]{45}{\textbf{MSA}} & \rotatebox[origin=c]{45}{\textbf{MER}} & 
\rotatebox[origin=c]{45}{\textbf{OV-MER}} & \rotatebox[origin=c]{45}{\textbf{ERI}} & 
\rotatebox[origin=c]{45}{\textbf{MIR}} & \rotatebox[origin=c]{45}{\textbf{ERG}} \\
\midrule
EmoLLMs \cite{liu2024emollms} & 7B & level 1 & \textcolor{OliveGreen}{\checkmark} & \textcolor{OliveGreen}{\checkmark} &  \textcolor{Maroon}{×} & \textcolor{Maroon}{×} & \textcolor{Maroon}{×} & \textcolor{Maroon}{×} \\
Emotion-LLaMA \cite{cheng2024emotion} & 7.8B & level 2 & \textcolor{Maroon}{×} & \textcolor{Maroon}{×} & \textcolor{OliveGreen}{\checkmark} & \textcolor{OliveGreen}{\checkmark} & \textcolor{Maroon}{×} & \textcolor{Maroon}{×} \\
Omni-Emotion \cite{yang2025omni} & 9B & level 2 & \textcolor{Maroon}{×} & \textcolor{Maroon}{×} & \textcolor{OliveGreen}{\checkmark} & \textcolor{OliveGreen}{\checkmark} & \textcolor{Maroon}{×} & \textcolor{Maroon}{×} \\
LGSRR \cite{zhou-etal-2025-llm} & 7.1B & level 2 & \textcolor{Maroon}{×} & \textcolor{Maroon}{×} & \textcolor{Maroon}{×} & \textcolor{Maroon}{×} & \textcolor{OliveGreen}{\checkmark} & \textcolor{Maroon}{×} \\
E3RG \cite{10.1145/3746027.3762029} & 7B & level 3 & \textcolor{Maroon}{×} &\textcolor{Maroon}{×} & \textcolor{Maroon}{×} & \textcolor{Maroon}{×} & \textcolor{Maroon}{×} & \textcolor{OliveGreen}{\checkmark} \\
EmoVIT \cite{xie2024emovit} & 8.2B & level 1\&2 & \textcolor{Maroon}{×} &\textcolor{OliveGreen}{\checkmark} & \textcolor{Maroon}{×} & \textcolor{OliveGreen}{\checkmark} & \textcolor{Maroon}{×} & \textcolor{Maroon}{×} \\
Empatheia \cite{zhang2025towards} & 8B & level 1\&3 & \textcolor{Maroon}{×} & \textcolor{OliveGreen}{\checkmark} & \textcolor{Maroon}{×} & \textcolor{Maroon}{×} & \textcolor{Maroon}{×} & \textcolor{OliveGreen}{\checkmark} \\
EmoVerse \cite{LI2025130810} & 4/8B & level 1\&2 & \textcolor{OliveGreen}{\checkmark} & \textcolor{OliveGreen}{\checkmark} & \textcolor{Maroon}{×} & \textcolor{OliveGreen}{\checkmark} & \textcolor{Maroon}{×} & \textcolor{Maroon}{×} \\
AffectGPT \cite{lian2025affectgpt} & 8.3B & level 1\&2 & \textcolor{OliveGreen}{\checkmark} & \textcolor{OliveGreen}{\checkmark} & \textcolor{OliveGreen}{\checkmark} & \textcolor{OliveGreen}{\checkmark} & \textcolor{Maroon}{×} & \textcolor{Maroon}{×} \\
Emotion-Qwen \cite{huang2025emotion} & 7.5B & level 1\&2 & \textcolor{Maroon}{×} & \textcolor{OliveGreen}{\checkmark} & \textcolor{Maroon}{×} & \textcolor{OliveGreen}{\checkmark} & \textcolor{Maroon}{×} & \textcolor{Maroon}{×} \\
SMES \cite{chu2025towards} & 7.1B & level 1\&3 & \textcolor{Maroon}{×} & \textcolor{OliveGreen}{\checkmark} & \textcolor{Maroon}{×} & \textcolor{Maroon}{×} & \textcolor{Maroon}{×} & \textcolor{OliveGreen}{\checkmark} \\
R1-Omni \cite{zhao2025r1} & 2.1B & level 1\&2 & \textcolor{Maroon}{×} & \textcolor{OliveGreen}{\checkmark} & \textcolor{Maroon}{×} & \textcolor{OliveGreen}{\checkmark} & \textcolor{Maroon}{×} & \textcolor{Maroon}{×} \\
\midrule
\textbf{Our Nano\mbox{-}EmoX} & 2.2B & level 1$ \sim $3 & \textcolor{OliveGreen}{\checkmark} & \textcolor{OliveGreen}{\checkmark} & \textcolor{OliveGreen}{\checkmark} & \textcolor{OliveGreen}{\checkmark} & \textcolor{OliveGreen}{\checkmark} & \textcolor{OliveGreen}{\checkmark} \\
\bottomrule
\end{tabular}
}
\end{table}


Nevertheless, this upward progression exposes a fundamental gap: current models are typically level specialists—they excel at tasks within a single cognitive stratum but fail to integrate knowledge across the hierarchy~\cite{liu2024emollms,cheng2024emotion,yang2025omni,zhou-etal-2025-llm,10.1145/3746027.3762029}. Developing a unified agent that spans the full perception-to-interaction continuum remains a major open challenge.
First, suboptimal fusion: existing fusion mechanisms struggle to adapt to the diverse feature requirements of different cognitive strata, limiting a model's ability to generalize across levels. Second, fragmented capabilities: as shown in \cref{tab:cap_compare}, the knowledge currently mastered by the model is still isolated. Without learning the deep connections between perceiving an emotion and reasoning about its cause, models lack genuine affective comprehension. Finally, resource intensity: the heavy computing and data demands of most LLM-based methods hinder the real-world deployment of a comprehensive affective agent for training and inference. Furthermore, deploying multiple task-specific models is impractical and inefficient.

To address these limitations, we introduce Nano\mbox{-}EmoX, a compact MLM that unifies six core affective tasks: multimodal sentiment analysis (MSA), multimodal emotion recognition (MER), open-vocabulary MER (OV\mbox{-}MER), multimodal intention recognition (MIR), emotion reason inference (ERI) and empathetic response generation (ERG). Specifically, Nano\mbox{-}EmoX integrates omni\mbox{-}modal inputs. Beyond capturing general visual and acoustic cues, our model explicitly models fine\mbox{-}grained facial affective signals and implements an early, hierarchical, and dynamic audio–visual feature fusion. After a dimensional alignment step performed by heterogeneous adapters, the language model (LM) proceeds to tackle all downstream tasks.

Building on the cognitive hierarchy, we propose a framework-P2E (Perception\mbox{-}to\mbox{-}Empathy), designed to efficiently unlock the model's potential in emotional intelligence. The core of this framework lies in a carefully designed data curriculum and a progressive training procedure. In itially, P2E enables the model to establish foundational perception and acquire multimodal fusion knowledge; subsequently, it cultivates advanced capabilities in affective reasoning and empathy. \cref{fig:task_hierarchy} illustrates our proposed conceptual hierarchy and summarizes the six unified affective tasks. Experimental results validate the effectiveness of the P2E framework in enabling learning across affective multilevel.

Our contributions are summarized as follows:
\begin{itemize}
    \item We present Nano\mbox{-}EmoX, a small-scale MLM that integrates a dedicated facial encoder and a hierarchical expert fusion encoder for dynamic audio–visual alignment. This design enables fine-grained affective feature modeling, and strong cross-task generalization across different cognitive levels.
    %
    %
    %
    %

    \item We introduce a three-level cognitive hierarchy that organizes affective tasks by their cognitive depth. Guided by this hierarchy, we then develop the P2E training framework, designed to progressively cultivate higher-level affective reasoning and emotional interaction.
    %
    %
    %
    %

    \item Nano\mbox{-}EmoX is the first compact MLM to unify six core affective tasks across all hierarchy levels, achieving comparable or better performance than substantially larger models. This demonstrates an effective balance between parameter efficiency and multilevel affective capability.
    %
    %
    %
    %

\end{itemize}


\begin{figure*}
  \centering
  \includegraphics[width=\textwidth]{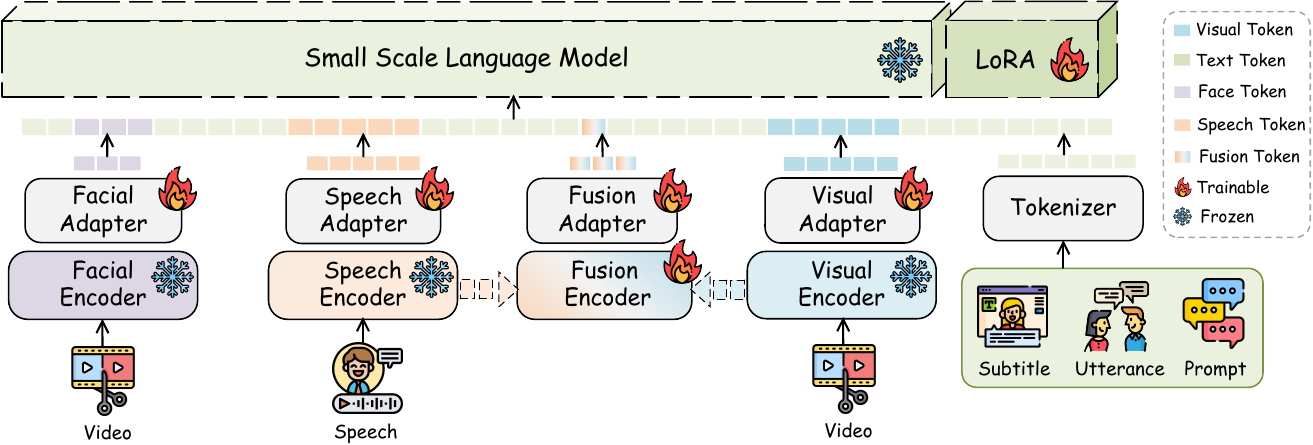}
  \caption{The architecture of the Nano\mbox{-}EmoX. The visual branch extracts general visual emotional cues, the facial branch is responsible for modeling fine\mbox{-}grained facial details, the speech branch captures acoustic emotional cues. To balance the contribution of each modality, the fusion branch integrates key emotional cues from the audio-visual modalities and extracts complementary information. The language model integrates multimodal information and performs multitask emotion recognition.}
  \label{fig:model overview}
\end{figure*}


\section{Related Work}
\label{sec:related_work} 


\paragraph{Multimodal Language Models.}
The advent of MLMs, \eg, SALMONN~\cite{tang2024salmonn}, Video-LLaMA~\cite{zhang-etal-2023-video}, Qwen2.5-Omni~\cite{xu2025qwen2} has revolutionized the field by integrating pre-trained modality encoders with LLMs. This fusion has led to remarkable advancements in tasks such as visual question answering and automatic speech recognition. Of late, comprehensive benchmarks \cite{sabour-etal-2024-emobench,zhang2025can,huemobench} have systematically evaluated both open-source and proprietary models, including InternLM 2.5~\cite{wu2024internlm2}, GPT-4~\cite{achiam2023gpt}, and Gemini-2.0-Flash~\cite{comanici2025gemini}, \etc. The findings consistently reveal a substantial emotional intelligence gap between MLMs and humans.

Recent research has increasingly focused on vertical MLMs for emotional domains. Works like EmoVIT~\cite{xie2024emovit} and Emotion-Qwen~\cite{huang2025emotion} focus on emotion recognition from vision, while subsequent studies such as Emotin-LLaMA~\cite{cheng2024emotion} and AffectGPT~\cite{lian2025affectgpt} incorporated audio and video features to achieve explainable emotion recognition. Other advancements have explored multimodal intent recognition tasks, as seen in \cite{MIntRec, zhou-etal-2025-llm}, or focused on generating more human-like empathetic responses~\cite{zhang2025towards,liu2024speak}. 

However, their work is limited by a lack of task-aware feature fusion and the absence of an explicit model for crucial facial expressions (\eg, AffecGPT). In contrast, Nano\mbox{-}EmoX employs adaptive fusion modeling and fine-grained facial-feature extraction, thereby boosting its multitask performance.


\paragraph{Multitask Learning for Emotion-centric MLMs.}
Recent works have seen a surge of interest in training paradigms for affective LLM-based mothod. For instance, EmoLLMs~\cite{liu2024emollms} utilizes instruction fine-tuning to optimize and unify five text-based sentiment tasks, while Emotion-LLaMA~\cite{cheng2024emotion} and AffectGPT~\cite{lian2025affectgpt} adopt joint-training to achieve a deeper understanding of emotion. Other approaches, Omni-Emotion~\cite{yang2025omni} employs multi-stage fine-tuning to enhance emotion processing capabilities. EmoVerse~\cite{LI2025130810}, have introduced M$^2$SE strategy to improve the emotional intelligence of MLMs. Furthermore, R1\mbox{-}Omni~\cite{zhao2025r1} leverages verifiable reinforcement learning to strengthen a model's emotional reasoning abilities. Nevertheless, most prior works remain confined to a single cognitive level. In contrast, our P2E training framework enables the model to learn capabilities that span the entire affective hierarchy—from perception to empathy.
\section{Methodology}
\label{sec:methodology}

\subsection{Architecture of Nano\mbox{-}EmoX}
\label{sec:nano_emox}
Nano\mbox{-}EmoX is a compact, hybrid-reasoning, and multitasking MLM designed for emotion-centric tasks. As depicted in \cref{fig:model overview}, it comprises four modality-specific branches and a language backbone.

\textbf{Scene visual perception branch:} To perceive generic visual signals, we employ a pre-trained visual encoder trained on large-scale datasets to process video frames $x_v \in \mathbb{R}^{3 \times H \times W}$, producing general-purpose visual emotion embeddings $E_v$. 
Here, $H$ and $W$ denote the frame height and width, respectively.

Since scaling down LMs typically reduces the number of modality-agnostic neurons~\cite{schwettmann2023multimodal}, we posit that incorporating a resampling network upstream of the LM can alleviate this problem and improve expressiveness. Therefore, the visual branch employs a two-layer Q-Former~\cite{li2023blip} to resample visual tokens $T_v \in \mathbb{R}^{T_v \times D_v}$, thereby enriching the emotional representation. In this notation, $T_v$ and $D_v$ represent the token length and dimensionality, respectively.

\textbf{Speech perception branch:} To extract high-quality acoustic features such as prosody and pitch, we employ a pre-trained speech encoder to process audio frames $x_a \in \mathbb{R}^{T_{a1} \times D_{a1}}$ (sampled at 1.6~kHz), obtaining speech emotion embeddings $E_a$. 
Here, $T_{a1}$ and $D_{a1}$ denote the frame length and the Mel-spectrogram dimensionality, respectively.

Similarly, the speech branch employs a two-layer Q-Former to extract fine-grained speech tokens $T_a \in \mathbb{R}^{T_{a2} \times D_{a2}}$, where $T_{a2}$ and $D_{a2}$ represent the token length and dimensionality, respectively.

\textbf{Enhanced face modeling:} Since facial expressions are crucial cues for conveying visual emotional features, modeling fine-grained facial representations is vital for enhancing the emotion perception capability of Nano\mbox{-}EmoX.

The FaceXFormer~\cite{Narayan_2025_ICCV} encoder excels at extracting fine-grained, identity-invariant facial representations. We enhance this encoder by shifting its processing paradigm from the original image-level operation to frame-sequence processing. Specifically, we employ a facial encoder to process video frames $x_v$ for the extraction of multi-scale features $E_f$. Subsequently, Temporal Modeling (TM) is responsible for reconstructing the temporal relationship of features, enabling the capture of key facial emotion expression $E_f^c$. The core computation of TM is formalized as follows:
\begin{equation}
    E_f^c = \text{CrossAttention}(Q, E_f^K, E_f^V)
    \label{eq:TAP}
\end{equation}
where $Q \in \mathbb{R}^{T_{f1} \times D_f}$ denotes learnable temporal query tokens, $T_{f1}$ is the  token length, $E_f^K$ and $E_f^V$ are the key and value projected from the face embedding $E_f$.

Subsequently, a two-layer fully connected network with GeLU~\cite{hendrycks2016gaussian} performs dimensional alignment with the LM and generates the face token $T_f \in \mathbb{R}^{T_f \times D_f}$, where $T_f$ and $D_f$ denote the token length and feature dimension, respectively. For detailed network specifications, Please refer to the appendix for more details.

\textbf{Cross-modal hierarchical expert fusion:} To enhance the model's multitask capabilities, we introduce a fusion encoder comprising three experts with independent weights and a gating network. Inspired by~\cite{zhao2024improving, chen2024finecliper, pastor2022cross, ding2023learning}—and recognizing the pivotal role of speech features in emotion-related tasks as well as the benefits of multiscale semantic information, we design a visual-speech fusion expert.
\begin{figure}[t]
    \centering
    \includegraphics[width=1.0\linewidth]{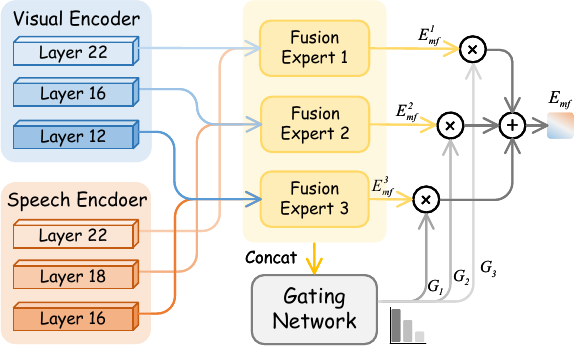}
    \caption{The fusion encoder extracts multi-layer features from the visual and speech encoders and feeds them to three fusion experts with independent weights. Each expert extracts complementary information $E_{mf}^i$. Then, the gating network dynamically weighs the contribution $G_{i}$ of each expert and routes the feature $E_{mf}$ of the output.}
    \label{fig:fusion_encoder_detail}
\end{figure}

The fusion expert employs speech features as queries to guide visual features through cross-modal cross-attention. Specifically, as described in \cref{fig:fusion_encoder_detail}, the encoder extracts intermediate features from layers 16, 18, and 22 of the speech encoder and from layers 12, 16, and 22 of the visual encoder. These features are hierarchically paired from lower to higher levels and fed into the three corresponding fusion experts to generate fused representations $E_{mf}^i$. The gating network, consisting of a two-layer fully connected network with GeLU, dynamically routes the output of each expert by adjusting their contributions $G_i$ based on the specific feature demands of each task. The gating network process is formulated as follows:
\begin{equation}
    G_1, G_2, G_3 = f_{\text{gate}}(\text{Concat}(E_{mf}^1, E_{mf}^2, E_{mf}^3))
    \label{eq:gate_network_weight}
\end{equation}
\begin{equation}
    E_{mf} = G_1 \odot E_{mf}^1 + G_2 \odot E_{mf}^2 + G_3 \odot E_{mf}^3
    \label{eq:gate_network_add_weight}
\end{equation}
Here, $E_{mf}$ represents the final fusion embedding and $f_{\text{gate}}$ denotes the gating network processing.

Finally, an adapter network projects the dimensionality of the fused features to align with that of the LM, generating fusion tokens $T_{mf} \in \mathbb{R}^{T_h \times D_h}$, where $T_{h}$ and $D_{h}$ denote the token length and feature dimension, respectively. With its hierarchical structure and dynamic gating mechanism, the fusion encoder effectively learns robust mappings between tasks and modalities. Please refer to the appendix for additional details.

\textbf{The core of language processing:} We use the Qwen2.5~\cite{qwen2025qwen25technicalreport} tokenizer to process dialogues, subtitles, and other text inputs to generate text tokens. The small-scale LM then integrates tokens from all modalities to accomplish various downstream emotion tasks.

\subsection{The P2E Training Framework}
\begin{figure*}
  \centering
  \includegraphics[width=\textwidth]{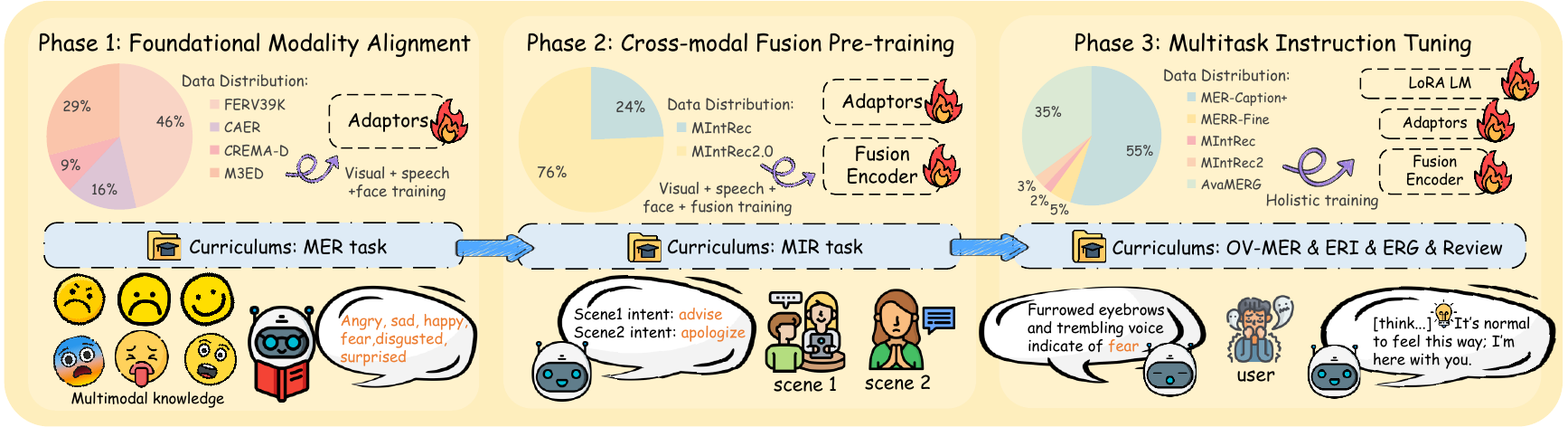}
  \caption{The P2E framework consists of a three-phase instruction fine-tuning process. Phase~1 focuses on the basic emotion recognition, to ensure a smooth learning curve, phase~2 multimodal fusion and contextual understanding by incorporating the MIR task. Finally, phase~3 revisits prior knowledge and integrates a diverse set of multilevel, complex tasks governed by a predefined data mixture ratio.}
  \label{fig:P2E}
\end{figure*}
As shown in \cref{fig:P2E}, the P2E framework is a three-phase curriculum, with each phase following the law of cognitive development from shallow to deep. We omit explicit training on the MSA task, as its knowledge can be implicitly acquired from related tasks. To adapt to different task formats, we use identifiers to distinguish tasks within the instruction templates, for the templates used in task training and additional details, please refer to the appendix.

The entire multitask training process is unified under a single objective: optimizing the maximum likelihood estimation (MLE) of the model parameters $\theta$ across all modalities, which can be formalized as follows:
\begin{equation}
    \theta^{MLE} = \arg\max_{\theta \in \Theta} \sum \log P\left(Y \mid T; \theta\right)
    \label{eq:MLE_optimization}
\end{equation}
where $T$ represents the tokenized representation of each modality, and $Y$ denotes the target output. The P2E framework achieves this objective through three carefully designed phases.

Phase 1: Foundational Modality Alignment. 
We first establish a robust unimodal foundation (see \cref{fig:P2E}, Phase~1). Training is focused on the modality-specific adapters to align the feature spaces of each encoder with the LM's embedding space, while the remaining modules are kept frozen. Specifically, the visual and facial adapters are jointly trained on FERV39K~\cite{wang2022ferv39k} and CAER~\cite{lee2019context} to learn diverse visual–language knowledge and fine-grained facial cues. Concurrently, the speech adapter is trained on CREMA-D~\cite{cao2014crema} and M3ED~\cite{zhao-etal-2022-m3ed} to capture emotional acoustic knowledge.

Phase 2: Cross-modal Fusion Pre-training. 
We posit that the MIR task serves as a natural bridge between basic perception and higher-order reasoning, compelling the model to synthesize multimodal cues to infer social goals (see \cref{fig:P2E}, Phase~2). 
We use the MIntRec~\cite{MIntRec} and MIntRec2.0~\cite{zhang2024mintrec} datasets to facilitate the learning of effective multimodal integration. 
This phase activates and trains the fusion encoder while continuing to train all modality adapters, whereas the remaining components are kept frozen. 
For models without a fusion-encoder architecture, this phase is retained to continue the joint training of adapters within each branch.

Phase 3: Multitask Instruction Tuning. 
This phase aims to cultivate synergy across tasks, from deepening fine-grained perception (OV-MER) to fostering high-level empathy (see \cref{fig:P2E}, Phase~3). 
We fine-tune the fusion encoder and all adapters on a carefully curated data curriculum and activate a Low-Rank Adaptation (LoRA) module for the LM. 
The data sampling ratio is set to MER: OV-MER: MIR: ERI: ERG = 18: 28: 5: 31: 18 to fully unleash the mode's potential.

For the OV-MER task, we use the MER-Caption+~\cite{lian2025affectgpt} dataset to train the model to capture fine-grained, multi-label emotional perception capabilities. For the ERI task, we employ MER-Caption+ together with the meticulously annotated MERR-Fine~\cite{cheng2024emotion} dataset to enhance the model's explanatory and reasoning abilities.

For the ERG task, we restructure the AvaMERG~\cite{zhang2025towards} dataset into a turn-based format, preserving dialogue history as context: \((Q_1^t, R_1^t), (Q_2^t, R_2^t), \dots\), where \(Q_i^t\) is the user's textual query and \(R_i^t\) is the model's previous textual response. 
In this format, the current user input \((Q_n^t, Q_n^v, Q_n^a)\) serves as the query, while the corresponding response \(R_n^t\) is the generation target. Here, \(Q_n^t\) represents the textual input, \(Q_n^v\) and \(Q_n^a\) denote the visual and audio queries, respectively, and \(R_n^t\) denotes the model's target textual response.


Following~\cite{zhang2025towards}, we guide the model to first consider the dialogue scenario, the speaker's emotion, and the response goal, thereby generating higher-quality empathetic responses. The reasoning process is wrapped within the \textless think\textgreater~tag. We also define a standard MER task on the AvaMERG dataset to help the model better grasp the associations between emotions and dialogue.
\section{Experimental Analysis}
\label{sec:experiment}
\begin{table*}[t]
\centering
\small  
\caption{Performance comparison on the MSA, MER and fine-grained OV-MER tasks. AffectGPT (s) marked with a $\dagger$ is trained solely on the MER-caption+ dataset, as proposed in the original work~\cite{lian2025affectgpt}, while the unmarked counterparts is trained using P2E. Nano\mbox{-}EmoX marked with $\ddagger$ uses a joint training approach. The best performance is displayed in \textbf{bold}, the second-best performance is \underline{underlined}.}
\label{tab:mer-unibench}
\renewcommand{\arraystretch}{1.05}
\resizebox{\linewidth}{!}{%
\begin{tabular}{@{\hspace{1pt}} c @{\hspace{2pt}} l@{\hspace{2pt}}|*{4}{c} @{\hspace{1pt}}|*{4}{c} @{\hspace{1pt}}|{c} @{\hspace{1pt}}|{c} @{\hspace{1pt}} }
\toprule
Models & Scale & MER2023~\cite{lian2023mer} & MER2024~\cite{lian2024mer} & MELD~\cite{he-etal-2019-towards} & IEMOCAP~\cite{busso2008iemocap} & MOSI~\cite{zadeh2016mosi}& MOSEI~\cite{zadeh2018multimodal}& SIMS~\cite{yu2020ch}& SIMSV2~\cite{liu2022make}& OV-MERD~\cite{lian2025ovmer}& Avg. \\
\midrule
& & \multicolumn{4}{c}{\cellcolor{purple!20}Hit Rate $\uparrow$ (MER)} & \multicolumn{4}{c}{\cellcolor{green!10}WAF $\uparrow$ (MSA)} & {\cellcolor{blue!10}WAF $\uparrow$ (OV-MER)}& - \\
\midrule
SALMONN~\cite{tang2024salmonn} & \textcolor{Maroon}{$\uparrow$ 11.7B} & 55.53 & 45.38 & 45.62 & 46.84 & 81.00 & 67.03 & 68.69 & 68.69 & 45.00 & 57.89 \\
MiniCPM-V-2.6-8B~\cite{hu2024minicpm} & \textcolor{Maroon}{$\uparrow$ 5.8B} & 46.67 & 45.31 & 40.27 & 36.31 & 74.96 & 57.44 & 74.85 & 75.04 & 50.04 & 55.65 \\
Qwen-2VL-7B~\cite{wang2024qwen2} & \textcolor{Maroon}{$\uparrow$ 5.5B} & 59.81 & 69.14 & 48.05 & 50.53 & 74.10 & 58.35 & 78.65 & 77.43 & 55.61 & 63.52 \\
Emotion-LLaMA~\cite{cheng2024emotion} & \textcolor{Maroon}{$\uparrow$ 5.6B} & 59.38 & 73.62 & 46.76 & 55.47 & 66.13 & 67.66 & 78.32 & 77.23 & 52.97 & 64.17 \\
AffectGPT~\cite{lian2025affectgpt} & \textcolor{Maroon}{$\uparrow$ 6.1B} & \underline{78.54} & \textbf{78.80} & \underline{55.65} & \textbf{60.54} & \textbf{81.30} & \textbf{80.90} & \textbf{88.49} & \textbf{86.18} & 62.52 & \textbf{74.77} \\
\midrule
\rowcolor{gray!30} \multicolumn{12}{c}{Small-scale Multimodal Models} \\
MobileVLM V2-1.7B~\cite{chu2024mobilevlm} & \textcolor{OliveGreen}{$\downarrow$ 0.5B} & 36.65 & 47.03 &  33.37 & 49.24 & 41.00 & 56.49 & 51.46 & 51.94 & 36.96 & 44.90 \\
MobileVLM V2-3B~\cite{chu2024mobilevlm} & \textcolor{Maroon}{$\uparrow$ 0.8B} & 37.70 & 53.87 &  30.72 & 53.00 & 55.81 & 44.87 & 69.17 & 65.72 & 33.86 & 49.95 \\
R1-Omni~\cite{zhao2025r1} & \textcolor{OliveGreen}{$\downarrow$ 0.1B} & 58.30 & 69.41 & 40.87 & 50.18 & 55.56 & 48.62 & 74.71 & 76.67 & 51.84 & 58.46 \\
AffectGPT (s)$^\dagger$ & \textcolor{OliveGreen}{$\downarrow$ 0.1B} & 73.45 & 74.71 & 47.69 & 53.14 & 75.51 & 71.30 & 82.50 & 84.10 & 62.43 & 69.43 \\
AffectGPT (s) & \textcolor{OliveGreen}{$\downarrow$ 0.1B} & 72.43 & 77.83 & 50.19 & 57.64 & 80.40 & \underline{79.97} & 83.28 & 83.23 & \underline{63.75} & 72.08 \\
\midrule
\textbf{Our Nano\mbox{-}EmoX$^\ddagger$} & 2.2B & 74.26 & \underline{78.61} & 54.27 & \underline{61.54} & \underline{80.71} & 79.52 & 84.64 & 83.31 & 62.68 & 73.28 \\
\textbf{Our Nano\mbox{-}EmoX} & 2.2B & \textbf{79.09} & 77.94 & \textbf{56.55} & 60.12 & 76.82 & 79.81 & \underline{86.25} & \underline{84.76} & \textbf{64.75} & \underline{74.01} \\
\bottomrule
\end{tabular}%
}
\end{table*}
In this section, we conduct a series of experiments to analyze the multitask processing capabilities of Nano\mbox{-}EmoX and to evaluate the performance gains achieved by the P2E framework. 
We perform evaluations on the following benchmarks: MER-UniBench~\cite{lian2025affectgpt}, EMER~\cite{lian2023explainable}, MIntRec~\cite{MIntRec}, MIntRec~2.0~\cite{zhang2024mintrec}, and AvaMERG~\cite{zhang2025towards}, where the evaluation metrics follow the official protocols. 
Please refer to the appendix for detailed descriptions of the benchmarks, evaluation metrics, and additional ablation experiments.

\subsection{Implementation Details}
\textbf{Model:} 
Nano\mbox{-}EmoX uses CLIP-Large~\cite{radford2021learning} as the visual encoder, HuBERT-Large~\cite{hsu2021hubert} as the speech encoder, and Qwen2.5-1.5B~\cite{qwen2025qwen25technicalreport} as the LM. 
The token lengths for the visual, speech, facial, and fusion streams are set to 32, 32, 4, and 1, respectively. 
For a controlled analysis of small-scale performance, we created AffectGPT(s), a compact variant of AffectGPT that substitutes the original LM with the smaller Qwen2.5-1.5B. This baseline was trained identically using our P2E framework.

\textbf{Training details:} 
We use AdamW~\cite{loshchilov2019decoupled} as the optimizer, with a batch size of 3 and gradient accumulation steps of 4. The model is trained on a single NVIDIA RTX 4090 GPU for 32 hours. The phase-specific hyperparameters for the P2E framework are as follows:  

Phase~1: The learning rate for all adapters is set to 3e-4. The visual and facial branches are trained for 25,000 steps, while the speech branch is trained for 15,000 steps.  
Phase~2: The learning rate for all trainable components is reduced to 1e-5, with training conducted for 5,000 steps.  
Phase~3: A uniform learning rate of 8e-6 is applied for 300,000 training steps. To conserve memory, the LoRA parameters are configured with $r = 32$ and $\alpha = 16$.

To evaluate our training methodology’s contribution, we compare the entire P2E with a standard Joint-training (Jo-T) setup, where all tasks are trained for 345000 steps jointly without hierarchical curriculum.
\begin{table}[t]  
\centering
\footnotesize
\caption{Performance comparison of models on the ERI task.}
\label{tab:emer}

\renewcommand{\arraystretch}{1.0}
\setlength{\tabcolsep}{2.3pt}
\begin{tabular}{c l c c}  
\toprule  
Models & Scale & Clue Overlap $\uparrow$ & Label Overlap $\uparrow$ \\
\midrule  
MiniCPM-V-2.6-8B~\cite{hu2024minicpm} & \textcolor{Maroon}{$\uparrow$ 5.8B} & 5.13 & 4.74 \\
Qwen-2VL-7B~\cite{wang2024qwen2} & \textcolor{Maroon}{$\uparrow$ 5.5B} & 6.32 & 5.65 \\
Emotion-LLaMA~\cite{cheng2024emotion} & \textcolor{Maroon}{$\uparrow$ 5.6B} & 7.83 & 6.25 \\
AffectGPT~\cite{lian2025affectgpt} & \textcolor{Maroon}{$\uparrow$ 6.1B} & 5.70 & 5.49 \\
Omni-Emotion~\cite{yang2025omni} & \textcolor{Maroon}{$\uparrow$ 6.8B} & \underline{8.22} & \underline{6.78} \\
Emotion-Qwen~\cite{huang2025emotion} & \textcolor{Maroon}{$\uparrow$ 5.3B} & \textbf{8.25} & \textbf{8.16} \\
\midrule 
\rowcolor{gray!30} \multicolumn{4}{c}{Small-scale Multimodal Models} \\ 
MobileVLM V2-1.7B~\cite{chu2024mobilevlm} & \textcolor{OliveGreen}{$\downarrow$ 0.5B} & 6.59 & 4.66 \\
MobileVLM V2-3B~\cite{chu2024mobilevlm} & \textcolor{Maroon}{$\uparrow$ 0.8B} & 6.49 & 4.82 \\
R1-Omni~\cite{zhao2025r1} & \textcolor{OliveGreen}{$\downarrow$ 0.1B} & 7.11 & 5.54 \\
AffectGPT (s) & \textcolor{OliveGreen}{$\downarrow$ 0.1B} & 7.60 & 5.70 \\
\midrule
\textbf{Our Nano\mbox{-}EmoX$^\ddagger$} & 2.2B & 7.62 & 5.46 \\
\textbf{Our Nano\mbox{-}EmoX} & 2.2B & 7.83 & 5.78 \\
\bottomrule 
\end{tabular}
\end{table}
\subsection{Performance Comparison}
\paragraph{Zero-shot Evaluation on The MSA, MER and OV-MER Task.}
To evaluate the intrinsic perception capabilities of the model developed under our P2E framework, we first assess its zero-shot performance on benchmarks without any task-specific fine-tuning. 
As shown in \cref{tab:mer-unibench}, Nano\mbox{-}EmoX achieves an overall average score that closely approaches current state-of-the-art (SOTA) results, achieving comparable results with 73\% fewer parameters.
Furthermore, our model establishes new SOTA results on both the coarse-grained emotion recognition benchmarks MER2023 and MELD, as well as on the OV-MER benchmark. Notably, AffectGPT (s) trained with our P2E framework exhibits a 3.6\% performance increase over the specialist AffectGPT (s), which was trained on its original method~\cite{lian2025affectgpt}. 

Our experiments validate both the proposed architecture and the training framework.  Specifically, Nano\mbox{-}EmoX demonstrates a strong ability to learn diverse emotional styles and generalize well, while the P2E framework is shown to improve the model's emotion awareness.

\paragraph{Zero-shot Evaluation on The ERI Task.}
As detailed in \cref{tab:emer}, our model surpasses numerous small-scale model and even larger-scale methods. 
Moreover, it performs on par with Emotion-LLaMA while using significantly fewer parameters. 
Nano\mbox{-}EmoX demonstrates strong proficiency in capturing subtle, context-dependent emotional cues within dynamic conversations, thereby validating the effectiveness of our architectural design for temporal reasoning and contextual understanding.
\paragraph{Fine-Tuning Evaluation on The MIR Task.}
As shown in \cref{tab:MIntrec}, Nano\mbox{-}EmoX achieves the best results among small-scale models, and surpasses a strong baseline, GPT\mbox{-}4, on MIntRec~2.0 by 12.4\% in accuracy (Acc) and 7.6\% in WF1. 
These results highlight the potential of Nano\mbox{-}EmoX in discriminating fine-grained intents.

In terms of performance, our method is not yet on par with substantially larger, SOTA models. However, our ablation study on visual tokens (detailed in the Appendix) demonstrates that increasing token counts effectively captures fine-grained emotional cues, significantly boosting MIR performance. This finding provides a clear direction for our future work on high-resolution affective modeling.
\begin{table}[t]
\centering
\footnotesize
\caption{Evaluation results of the model on the MIR task.}
\label{tab:MIntrec}

\renewcommand{\arraystretch}{1.05}
\setlength{\tabcolsep}{2.1pt}

\begin{tabular}{c c c c c c c}
\toprule  
Methods & \multicolumn{3}{c}{MIntRec} & \multicolumn{3}{c}{MIntRec 2.0} \\  
\cmidrule(lr){2-4} \cmidrule(lr){5-7}  
& Acc $\uparrow$ & WF1 $\uparrow$& WP $\uparrow$ & Acc $\uparrow$ & WF1 $\uparrow$ & WP $\uparrow$ \\ 
\midrule  
GPT-4 (3 shots)~\cite{achiam2023gpt} & 63.84 & 63.97 & 68.16 & 42.10 & 43.92 & 53.18 \\   
MiniCPM-V-2.6-8B~\cite{hu2024minicpm} & 80.67 & 80.56 & 82.19 & 53.58 & 51.91 & 61.66 \\  
Qwen-2VL-7B~\cite{wang2024qwen2} & \underline{82.92} & \underline{82.79} & \underline{86.75} & 64.19 & 63.31 & 64.39 \\ 
Qwen-VL-72B~\cite{wang2024qwen2} & \textbf{86.29} & \textbf{86.09} & \textbf{86.75} & \textbf{66.99} & \textbf{66.63} & \textbf{67.45}\\
LLaVA-Video-72B~\cite{lin2023video} & 80.22 & 79.94 & 80.63 & \underline{64.98} & \underline{64.72} & \underline{65.40} \\
\midrule 
\rowcolor{gray!30} \multicolumn{7}{c}{Small-scale Multimodal Models} \\ 
MobileVLM V2-1.7B~\cite{chu2024mobilevlm} & 33.27 & 33.81 & 34.75 & 21.34 & 22.64 & 25.59 \\  
MobileVLM V2-3B~\cite{chu2024mobilevlm} & 36.12 & 37.74 & 35.12 & 23.98 & 24.16 & 26.45 \\    
AffectGPT (s) & 48.76 & 50.71 & 57.87 & 38.79 & 37.18 & 41.91 \\  
\midrule 
\textbf{Our Nano\mbox{-}EmoX$^\ddagger$} & 56.18 & 56.62 & 65.05 & 47.13 & 45.38 & 47.83 \\  
\textbf{Our Nano\mbox{-}EmoX} & 58.20 & 58.17 & 60.12 & 47.32 & 47.27 & 51.10 \\  
\bottomrule  
\end{tabular}

\setlength{\tabcolsep}{6pt}
\end{table}
\begin{table}[t]
\centering
\footnotesize  
\caption{Performance comparison of models with different scales on the ERG task. MiniCPM-o-2.6B~\cite{hu2024minicpm}, Qwen2.5-Omni-7B~\cite{xu2025qwen2}, and Ola-Omni-7B~\cite{liu2025ola} enhance their capabilities by incorporating the E3RG~\cite{10.1145/3746027.3762029} method.}
\label{tab:Avamerg}

\renewcommand{\arraystretch}{1.05}
\setlength{\tabcolsep}{5.8pt}

\begin{tabular}{c c c c}
\toprule  
Models & Acc / Hit Rate $\uparrow$ & Dist-1 $\uparrow$ & Dist-2 $\uparrow$ \\  
\midrule  
Empatheia~\cite{zhang2025towards} & 48.51 / - & 2.69 & 14.76 \\  
MiniCPM-o-2.6B~\cite{hu2024minicpm} & - / 65.8 & 95.20 & 99.60 \\  
Qwen2.5-Omni-7B~\cite{xu2025qwen2} & - / 72.3 & \underline{98.60} & 99.70 \\  
Ola-Omni-7B~\cite{liu2025ola} & - / 75.6 & \textbf{98.90} & 99.80 \\  
\midrule  
\rowcolor{gray!30} \multicolumn{4}{c}{Small-scale Multimodal Models} \\
MobileVLM V2-1.7B~\cite{chu2024mobilevlm} & 4.07 / 21.14 & 75.07 & 96.91 \\
MobileVLM V2-3B~\cite{chu2024mobilevlm} & 25.18 / 61.42 & 78.66 & 97.41 \\ 
AffectGPT (s) & \underline{47.36} / \underline{90.88} & 95.44 & \underline{99.81} \\ 
\midrule
\textbf{Our Nano\mbox{-}EmoX$^\ddagger$} & 34.91 / 74.79 & 95.41 & 99.73 \\  
\textbf{Our Nano\mbox{-}EmoX} & \textbf{54.82 / 91.13} & 95.47 & \textbf{99.83} \\  
\bottomrule  
\end{tabular}

\setlength{\tabcolsep}{6pt}
\end{table}
\paragraph{Fine-Tuning Evaluation on The ERG Task.}
As detailed in \cref{tab:Avamerg}, our model continues to demonstrate strong performance. 
Nano\mbox{-}EmoX achieves an Acc of 54.82 in recognizing the speaker's emotion prior to generating an empathetic response, surpassing all other small-scale models. 
Notably, it establishes a new SOTA in coarse-grained emotion recognition with a Hit Rate of 91.13, marking a 22.54\% improvement over the previous SOTA method. 
Meanwhile, AffectGPT (s), after being trained with the P2E framework, achieves the second-best performance with a Hit Rate of 90.88. 
These results validate the effectiveness of our cross-modal fusion encoder and the P2E framework for high-quality empathetic generation, highlighting our approach as a promising solution for enabling models to master rich, multi-layered emotional knowledge.

\subsection{Ablation Evaluation}
\paragraph{Investigating Performance Improvements of The P2E Framework on Affective Tasks.}
The results, presented in \cref{tab:ablation_training}, are striking. 
While the standard Jo-T approach provides a solid foundation, particularly for perceptual tasks within MER-UniBench, our strategy consistently outperforms it across all task categories. 
The most pronounced improvements are observed in the ERG task. After P2E training, the Hit Rate of AffectGPT (s) increases by an impressive 67.72\%, while that of Nano\mbox{-}EmoX rises by 17.93\%. 
These findings strongly corroborate the effectiveness of the P2E framework in deepening emotional perception and fostering a more comprehensive form of emotional intelligence.

To further validate our approach, we then reverse the shallow-to-deep cognitive progression by designing a "Reverse P2E" training strategy. As shown in \cref{tab:reversep2e_ablation}, reverse P2E leads to a noticeable performance degradation. This results underscore the importance of adhering to a cognitively-aware, shallow-to-deep learning paradigm for cultivating the model's affective capabilities.
\begin{table}[t]
\centering
\footnotesize
\caption{Comparison of training time and performance improvements across different training methods, evaluated on the MER-UniBench~\cite{lian2025affectgpt}, EMER~\cite{lian2023explainable}, and AvaMERG~\cite{zhang2025towards} benchmarks.}
\label{tab:ablation_training}

\setlength{\tabcolsep}{0.5pt}

\begin{tabular}{c c c c c c}
\toprule  
\multirow{2}{*}{Models} & \multirow{2}{*}{Strategy} & \multirow{2}{*}{Time} & MSA\&MER\&OV-MER & ERI & ERG \\
\cmidrule{4-6}  
& & & Avg. $\uparrow$ & Avg. $\uparrow$ & Hit Rate $\uparrow$ \\
\midrule  
AffectGPT (s) & Jo-T & 40 h & 72.26 & 6.71 & 29.33 \\
Our Nano\mbox{-}EmoX & Jo-T & 40 h &\underline{73.28} & 6.54 & 74.79 \\
AffectGPT (s) & P2E &32 h& 72.08 & \underline{6.65} & \underline{90.88} \\
Our Nano\mbox{-}EmoX & P2E &32 h& \textbf{74.01} & \textbf{6.80} & \textbf{91.13} \\
\bottomrule  
\end{tabular}

\setlength{\tabcolsep}{6pt}
\end{table}
\begin{table}[t]
\centering
\footnotesize
\caption{The comparison result between the standard P2E and the reverse P2E (from empathy to perception).}
\label{tab:reversep2e_ablation}
\setlength{\tabcolsep}{3pt}
\begin{tabular}{c c c c}
\toprule
\multirow{2}{*}{Strategy} & MSA \& MER \& OV-MER & ERI & ERG \\
\cmidrule{2-4}
 & Avg. $\uparrow$ & Avg. $\uparrow$ & Hit Rate $\uparrow$ \\
\midrule
Standard P2E & 74.01 & 6.80 & 91.13 \\
Reverse P2E &63.35 \textcolor{Maroon}{(-10.66)} & 6.17 \textcolor{Maroon}{(-0.63)} & 57.64 \textcolor{Maroon}{(-33.49)} \\
\bottomrule
\end{tabular}

\setlength{\tabcolsep}{6pt}
\end{table}
\begin{figure*}[t] 
    \centering
    \includegraphics[width=1.0\textwidth]{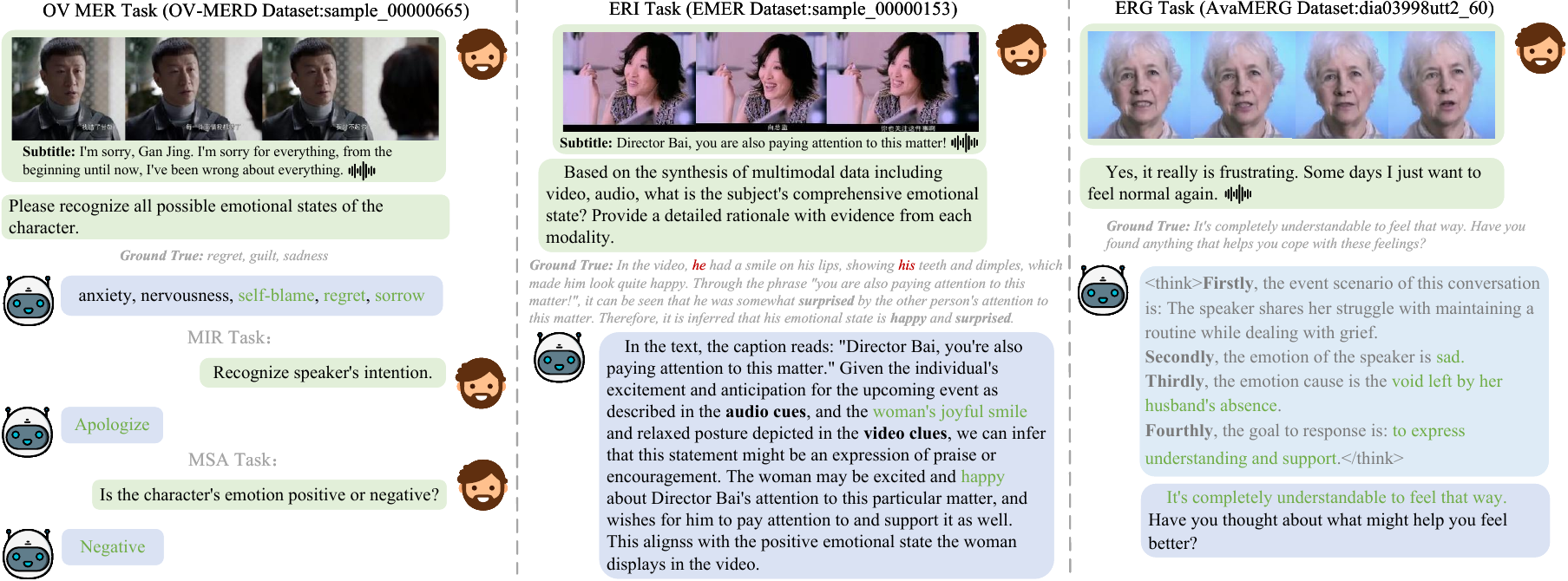}
    \caption{Visualization results on the OV-MER, MIR, ERI, and ERG tasks. Our model responds rapidly when handling perceptual tasks, whereas for interactive tasks, it first considers the interlocutor's state before generating empathetic responses.}
    \label{fig:qualitative_analysis}
\end{figure*}
\paragraph{Revisiting the Role of MSA Task Training.}
\cref{fig:3key_ablation} (a) presents the performance of Nano\mbox{-}EmoX on the MSA task under both fine-tuned and zero-shot settings. Notably, the model maintains competitive accuracy even without task-specific fine-tuning. We attribute this strong zero-shot capability to the presence of positive and negative annotations in our pre-training data, from which the model implicitly acquired knowledge relevant to the MSA task.

\paragraph{Architectural Contribution Analysis.}
As detailed in \cref{tab:modality_contribution}, the bimodal variants serve as strong baselines but struggle on the complex ERG task, underscoring the need for a more holistic perceptual system. 
Augmenting the visual representations with our facial encoder significantly improves performance, which validates the importance of leveraging facial cues for effective affective modeling.

The effectiveness of our specific design choices is further validated by two key experiments. 
First, when we replace our facial encoder with face landmarks and Action Units (AUs) extracted using FAN~\cite{bulat2017far} and ME\mbox{-}GraphAU~\cite{ijcai2022p173} respectively, the model's performance declines.
Second, removing the fusion encoder entirely or using attention fusion~\cite{lian2025affectgpt} leads to a significant decline in performance on AvaMERG, clearly demonstrating its effectiveness in dynamically and efficiently integrating multimodal streams.

We also explored alternative fusion architectures to validate the effectiveness of our design. Specifically, replacing the layer-wise sequential fusion with a cross-layer approach, or substituting the dynamic gating mechanism with average weighting, both resulted in consistently lower performance. As shown in \cref{fig:3key_ablation} (b) and \cref{fig:3key_ablation} (c), our encoder effectively learns to fuse features adaptively, selecting combinations that are better suited for each task.
    %
    %


    %
    %

\begin{table}[t]
\centering
\footnotesize
\caption{Ablation results showing the influence of different branch combinations in Nano\mbox{-}EmoX, `A', `V', `F', `M', and `T' represent Audio, Vision, Face, Fusion, and Text modalities, respectively.}
\label{tab:modality_contribution}
\setlength{\tabcolsep}{0.09pt}

\begin{tabular}{c c c c c}
\toprule
\multirow{2}{*}{Variants} & \multirow{2}{*}{Modal} & MSA\&MER\&OV-MER & ERI & ERG \\
\cmidrule{3-5}
& & Avg. $\uparrow$ & Avg. $\uparrow$ & Hit Rate $\uparrow$ \\
\midrule  
Audio Perception         & AT    & 69.23 & 5.86 & 76.50 \\
Visual Perception       & VT    & 64.13 & 5.78 & 53.36 \\
Face Perception          & FT    & 65.99 & 5.63 & 46.54 \\
Visual + Face  & VFT   & 67.26 & 5.94 & 82.61 \\
w/o Fusion& AVFT  & \underline{71.65} & 6.05 & 62.20 \\
w  Experts Fusion & AVFMT & \textbf{74.01} & \textbf{6.80} & \textbf{91.13} \\
Face Landmark + AU     & AVFMT & 70.88 & 6.13 & 57.25 \\
Attention Fusion     & AVFMT & 71.43 & \underline{6.56} & \underline{86.27} \\
\bottomrule  
\end{tabular}
\setlength{\tabcolsep}{6pt} 

\end{table}
\begin{figure}[t] 
    \centering
    \includegraphics[width=1.0\linewidth]{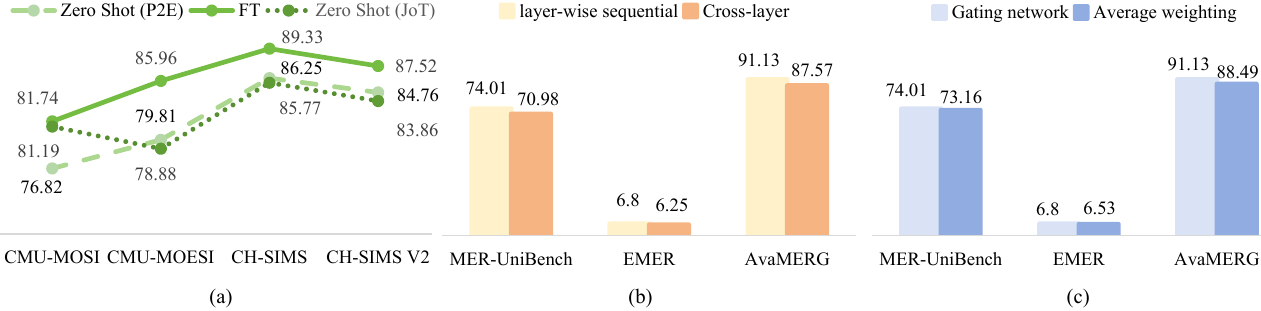} 
    \caption{Ablation result: (a) Training the MSA task vs. Zero shot; (b) Feature layer-wise fusion vs. cross-layer fusion; (c) Gated dynamic weighting vs. average weighting.}
    \label{fig:3key_ablation}
\end{figure}

\paragraph{Qualitative analysis on Nano\mbox{-}EmoX.}
We present visualization results of Nano\mbox{-}EmoX on four key tasks in the \cref{fig:qualitative_analysis}. In the OV-MER task, our model accurately captures more subtle emotions such as tension and anxiety. 
In the ERI task, by comprehensively synthesizing multimodal emotional cues, Nano\mbox{-}EmoX achieves precise causal reasoning for a character's emotional state.
Furthermore, in the ERG task, Leveraging a chain-of-thought process, Nano\mbox{-}EmoX first analyzes the emotional context by pinpointing the user's emotion and its root cause. It then establishes a response objective to guide the generation of a final, empathetic, and contextually appropriate reply. For more visualization results, please refer to the appendix.

\section{Conclusion}
\label{sec:conclusion}
In this work, we propose a three-level conceptual hierarchy that mirrors the cognitive progression, building on this taxonomy, we introduce Nano\mbox{-}EmoX, a compact MLM, and the P2E training framework, which together resolve the long-standing fragmentation of skills in affective computing. Nano\mbox{-}EmoX improves cross-task generalization via fine-grained facial modeling and audio-visual dynamic fusion, while P2E validates that cognition-inspired, progressive curricula are indispensable for cultivating multilevel emotional intelligence. Our results show that advancing emotional AI hinges less on scaling model size than on co-designing efficient architectures with structured, cognition-aligned training. 

\clearpage
\appendix
\setcounter{page}{1}
\setcounter{table}{0}
\setcounter{figure}{0}
\setcounter{section}{0}
\setcounter{equation}{0}
\renewcommand{\thesection}{\Alph{section}}

\renewcommand{\theHsection}{appendix.\thesection} 

\maketitlesupplementary
\section{Overview}
As part of the Appendix, we present the following as an extension to the ones shown in the paper:

\begin{itemize}[leftmargin=*, label=$\bullet$] 
    \item Task Definition (\cref{sup:task define})
    \item Nano\mbox{-}EmoX Details (\cref{sup:modeldetials})
    \item Details of P2E Framework (\cref{sup:P2Edetails})
    \item Experimental Setup and Additional Experiments (\cref{sup:details_of_exp})
    \item More visualization results (\cref{sup:more_visual_res})
\end{itemize}

\section{Task Definition}
\label{sup:task define}
The P2E is conceptually inspired by Preston \& de Waal’s PAM. We map this to P2E as (1) perception: non-deliberative extraction of affective cues from multimodal inputs (automatic activation), (2) understanding: context and intent-aware integration (regulatory modulation), and (3) interaction: generation of context-appropriate, socially aligned outputs (prosocial response). 

\paragraph{Level 1: Foundational Perception.}
\emph{Multimodal Sentimental Analysis (MSA):} This task takes as input multimodal data including text, images, and speech. It fuses emotion-related features across these modalities—such as textual semantics, facial expressions in images, and prosody in speech—and determines the emotional state of the target. The emotional state can be categorized by sentiment polarity (positive/negative/neutral) or emotional intensity levels.

\emph{Multimodal Emotion Recognition (MER):} This task involves identifying discrete emotion categories (\eg, joy, sadness) or continuous affective dimensions from human expressions. 

\emph{Open-Vocabulary MER (OV-MER): }Moving beyond coarse-grained labels, OV-MER requires the model to identify and describe nuanced, intertwined emotions (\eg, a mix of anxiety and anger). 

\paragraph{Level 2: Deep Understanding.}
\emph{Emotion Reasoning Integration (ERI):} This task pushes the model beyond mere recognition into the realm of causal inference, requiring it to explain the underlying reasons for a specific emotion. 

\emph{Multimodal Intent Recognition (MIR):} To understand the social goals behind utterances, MIR requires the model to infer a speaker's intent (\eg, gratitude, suggestion, apology) from both verbal and non-verbal cues. 

\paragraph{Level 3: Emotional Interaction.}
\emph{Empathic Response Generation (ERG):} This task takes as input the user’s emotional expressions (e.g., text, speech) and contextual information. It first understands the user’s emotional needs and underlying emotions, then generates natural language responses that align with the user’s emotions and convey understanding and support, ultimately achieving emotional resonance.

\section{Details of Nano\mbox{-}EmoX}
\label{sup:modeldetials}
\begin{figure}[t]
    \centering
    \includegraphics[width=1.0\linewidth]{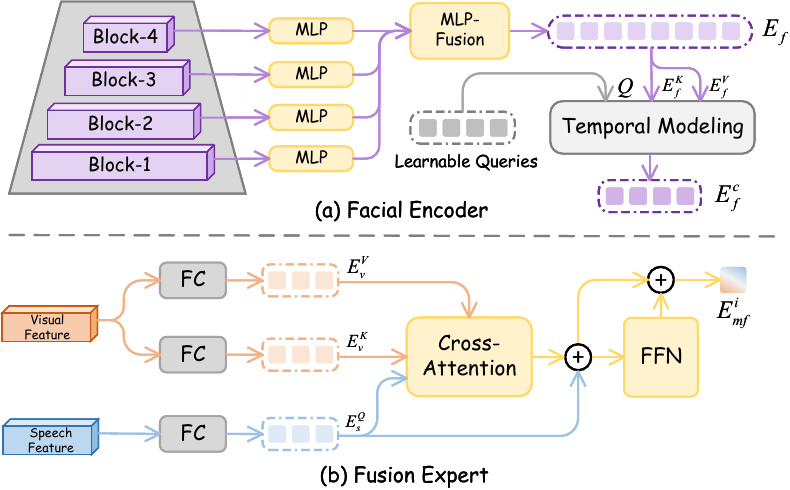}
    \caption{The facial Encoder extracts multiscale facial features and fuses them via an MLP to generate a rich facial embedding $E_{f}$. Subsequently, a temporal modeling block construct the sequence to output a final facial representation, which provides the language model with critical affective visual signals $E_{f}^c$. Fusion experts use audio features to guide vision and extract key complementary information $E_{mf}^i$.}
  \label{fig:model detail}
\end{figure}

\subsection{Fine-grained Facial Clues Extracting}
\label{sup:Facialencoder details}
The \cref{fig:model detail} (a) illustrates the network details: a lightweight facial encoder extracts features from block-1, block-2, block-3, and block-4 of the visual backbone network, which encompasses multiscale facial features ranging from fine-grained to global semantics. Features at each scale are aligned and then aggregated into MLP Fusion, which fuses them into a unified representation balancing facial detail and global structure:
\begin{equation}
    E_f = f_{\text{FaceXFormer}}(x_v)
    \label{eq:facial_encoding}
\end{equation}

$\quad E_f \in \mathbb{R}^{T_f \times D_f}$, where $T_f$ and $D_f$ denote the length and dimension of embeddings, respectively. To extend the facial encoder's capability from single-frame to video-level analysis, we introduce learnable temporal query tokens $Q$. These tokens interact with frame-ordered facial features via temporal modeling to reconstruct the time-sequential relationships among key facial emotional cues. The specific calculation methods and subsequent processing steps are presented in Sec. 3.1.



\subsection{Fusion Expert}
\label{sup:Fusionexpert_details}
The details of fusion expert as depicted in~\cref{fig:model detail} (b), The fusion process within each expert $i$ is formalized as:
\begin{equation}
    E_{m}^i = \text{CrossAttention}(E_s^Q, E_v^K, E_v^V) + E_s^Q
    \label{eq:fusion expert ca}
\end{equation}

where $E_s^Q$ denotes the query features projected from the speech embedding $E_s$, and $E_v^K$ and $E_v^V$ represent the key and value projections from the visual embedding $E_v$. This allows the fusion expert to leverage the more emotionally stable speech cues to attend to the most salient affective information within the visual stream. Subsequently, a feed-forward network (FFN) enriches the representation:
\begin{equation}
    E_{mf}^i = \text{FFN}(E_{m}^i) + E_{m}^i
    \label{eq:fusion expert ffn}
\end{equation}




\section{Details of P2E Framework}
\label{sup:P2Edetails}
\begin{table*}[t]
\centering
\caption{Details of task identifiers and training datasets for diverse emotional tasks. }
\label{tab:emotion_task_details}
\footnotesize 
\begin{tabular}{lccccc}
\toprule
Task       & MER                     & OV-MER         & ERI                             & MIR                 & ERG        \\
\midrule
Identifier & {[}Recognition{]}       & {[}Recog\_OV{]} & {[}Inference{]}                 & {[}Intent{]}        & {[}Interaction{]} \\
Datasets   & CAER~\cite{lee2019context}, CREMA-D~\cite{cao2014crema}         & MER-Caption+~\cite{lian2025affectgpt}   & MER-Caption+~\cite{lian2025affectgpt}                    & MIntRec~\cite{MIntRec}        & AvaMERG~\cite{zhang2025towards}    \\
         & M3ED~\cite{zhao-etal-2022-m3ed}, FERV39K~\cite{wang2022ferv39k}             &                & MER-Fine~\cite{cheng2024emotion}                        & MIntRec2.0~\cite{zhang2024mintrec}                      &         \\
Samples    & 141k                    & 36k            & 40.5k                           & 7.4k                & 57k        \\
\bottomrule
\end{tabular}
\end{table*}
In this section, we provide specific additional details about the P2E framework, including the prompt templates used for training. \cref{tab:emotion_task_details} describes the task identifiers and training data used for each training task. 

Task identifiers are essential for the model to accurately follow instructions. Embedded within the P2E curriculum, these identifiers enable the model to execute rapid reasoning in perception and understanding layers, and employ Chain-of-Thought for deep contemplation in the interactive empathy layer, thereby ensuring the output of accurate and appropriate empathetic responses
\paragraph{Phase1: Foundational Modality Alignment:} in this initial stage (see Fig.4, Phase 1 in the Sec 3.2.), we focus on pre-training for basic emotion recognition to establish a robust foundation by aligning the feature space of each modality encoder with the language model's embedding space. 
An example of the standardized instruction template for this phase is shown below:

\begin{tcolorbox}[
    colback=gray!20,       
    colframe=black,        
    width=\linewidth,      
    boxsep=0pt,            
    left=0pt, right=0pt,   
    top=0pt, bottom=0pt,   
    before skip=1em,
    after skip=1em
]
\colorbox{black}{%
    \makebox[\dimexpr\linewidth-2\fboxsep\relax][l]{  
    \hspace{10pt}
        \textcolor{white}{\bfseries The MER Task Prompt Template}%
    }%
}

\vspace{6pt} 
\begin{minipage}{\linewidth}
    \hspace{10pt}
    \begin{minipage}{\dimexpr\linewidth-20pt}
        \textbf[Recognition] Please select the label that can best describe the person's emotional state from the provided candidate labels: \textless Emotion Labels\textgreater.
    \end{minipage}
\end{minipage}
\vspace{6pt} 
\end{tcolorbox}

\paragraph{Phase2: Cross-modal Fusion Pre-training:} We posit that intent recognition serves as a natural bridge between basic perception and higher-order empathy, as it requires the model to synthesize cross-modal cues to infer a speaker's underlying social goals, a clear progression from simple emotion identification.
The instruction template for the MIR task is as follows:

\begin{tcolorbox}[
    colback=gray!20,       
    colframe=black,        
    width=\linewidth,      
    boxsep=0pt,            
    left=0pt, right=0pt,   
    top=0pt, bottom=0pt,   
    before skip=1em,
    after skip=1em
]
\colorbox{black}{%
    \makebox[\dimexpr\linewidth-2\fboxsep\relax][l]{  
    \hspace{10pt}
        \textcolor{white}{\bfseries The MIR Task Prompt Template}%
    }%
}

\vspace{6pt} 
\begin{minipage}{\linewidth}
    \hspace{10pt}
    \begin{minipage}{\dimexpr\linewidth-20pt}
        \textbf[Intent] Recognize speaker's intention from the provided candidate labels: \textless Intention Labels\textgreater.
    \end{minipage}
\end{minipage}
\vspace{6pt} 
\end{tcolorbox}

\paragraph{Phase3: Multitask Instruction Tuning:} in the final stage (see Fig.4, Phase 3 in the Sec 3.2.), we fine-tune the entire architecture on a complex mixture of tasks to integrate all acquired knowledge and unlock the model's full potential for high-level reasoning and empathetic interaction. 

\textbf{Deepening perception:} to facilitate the model in learning to address the OV-MER task, which requires describing fine-grained and multi-label emotions, we have specified the following prompt template:
\begin{tcolorbox}[
    colback=gray!20,       
    colframe=black,        
    width=\linewidth,      
    boxsep=0pt,            
    left=0pt, right=0pt,   
    top=0pt, bottom=0pt,   
    before skip=1em,
    after skip=1em
]
\colorbox{black}{%
    \makebox[\dimexpr\linewidth-2\fboxsep\relax][l]{  
    \hspace{10pt}
        \textcolor{white}{\bfseries The OV-MER Task Prompt Template}%
    }%
}

\vspace{6pt} 
\begin{minipage}{\linewidth}
    \hspace{10pt}
    \begin{minipage}{\dimexpr\linewidth-20pt}
        \textbf[Recogn\_OV] Recognize all the possible emotional states the character miaght be feeling in this context.
    \end{minipage}
\end{minipage}
\vspace{6pt} 
\end{tcolorbox}

\textbf{Cultivating Reasoning:} for the ERI task, we require the model to describe the most relevant emotional cues, with the prompt template as follows:
\begin{tcolorbox}[
    colback=gray!20,       
    colframe=black,        
    width=\linewidth,      
    boxsep=0pt,            
    left=0pt, right=0pt,   
    top=0pt, bottom=0pt,   
    before skip=1em,
    after skip=1em
]
\colorbox{black}{%
    \makebox[\dimexpr\linewidth-2\fboxsep\relax][l]{  
    \hspace{10pt}
        \textcolor{white}{\bfseries The ERI Task Prompt Template}%
    }%
}

\vspace{6pt} 
\begin{minipage}{\linewidth}
    \hspace{10pt}
    \begin{minipage}{\dimexpr\linewidth-20pt}
        \textbf[Inference] From the combined evidence of speech, tone, and visual expression, construct a detailed summary of the subject's emotional journey and final inferred state.
    \end{minipage}
\end{minipage}
\vspace{6pt} 
\end{tcolorbox}

\begin{tcolorbox}[
    colback=gray!20,       
    colframe=black,        
    width=\linewidth,      
    boxsep=0pt,            
    left=0pt, right=0pt,   
    top=0pt, bottom=0pt,   
    before skip=1em,
    after skip=1em
]
\colorbox{black}{%
    \makebox[\dimexpr\linewidth-2\fboxsep\relax][l]{  
    \hspace{10pt}
        \textcolor{white}{\bfseries The ERG Task Prompt Template}%
    }%
}

\vspace{6pt} 
\begin{minipage}{\linewidth}
    \hspace{10pt}
    \begin{minipage}{\dimexpr\linewidth-20pt}
        \textbf[Interaction] You are an empathetic listener, your goal is to understand the user’s emotions and intentions, and respond or comfort them with appropriate language that helps them feel understood and cared for. Please analyze using Chain of Empathy:
        
\textbf{First,} Reflect on the event scenarios that arise from the ongoing dialogue.

\textbf{Secondly,} Analyze both the implicit and explicit emotions conveyed by the user.

\textbf{Thirdly}, Infer the underlying reasons for the user's emotions.

\textbf{Fourthly}, Determine the goal of your response in this particular instance, such as alleviating anxiety, offering reassurance, or expressing understanding.
    \end{minipage}
\end{minipage}
\vspace{6pt} 
\end{tcolorbox}

\textbf{Empathy activation:} to enable the model to generate the most appropriate empathetic responses based on prior knowledge, we require it to engage in step-by-step reasoning following a four-step approach. After this deliberative empathetic process, the model then generates the final response to the interlocutor. The ERG task prompt template is illustrated above.

\section{Experimental Setup and Additional Experiments}
\label{sup:details_of_exp}
\subsection{Benchmarks}
\label{sup:benchmarks}
Our comprehensive evaluation assesses performance across six core affective tasks using a suite of established benchmarks. A significant portion of this evaluation is conducted using MER-UniBench~\cite{lian2025affectgpt}, a multifaceted benchmark designed for three distinct tasks:

The MSA task is evaluated on the standard benchmarks of MOSI (CMU-MOSI)~\cite{zadeh2016mosi}, MOSEI (CMU-MOSEI)~\cite{zadeh2018multimodal}, SIMS (CH-SIMS)~\cite{yu2020ch}, and its successor, SIMSv2 (CH-SIMS V2)~\cite{liu2022make}.

The MER task is assessed on subsets of four widely-used datasets: MER2023~\cite{lian2023mer}, MER2024~\cite{lian2024mer}, MELD~\cite{he-etal-2019-towards}, and IEMOCAP~\cite{busso2008iemocap}.
The OV-MER task is benchmarked against the specialized OV-MERD~\cite{lian2025ovmer} dataset.

For the remaining three affective tasks, we employ the following four benchmarks:

The explainable ERI task is evaluated using the primary EMER~\cite{lian2023explainable} benchmark.
The MIR task is assessed on the standard MIntRec~\cite{MIntRec} and MIntRec2.0~\cite{zhang2024mintrec} testset.

The ERG task utilizes the large-scale AvaMERG~\cite{zhang2025towards} testset for evaluation.

\subsection{Metrics}
\label{sup:metrics}
To ensure fair and comprehensive comparisons, we adopt the official evaluation metrics for each benchmark.
\begin{enumerate}
    \item For the MER task, following MER-UniBench~\cite{lian2025affectgpt}, we report the Emotion Wheel Hit Rate. This metric provides a robust measure of categorical accuracy by mapping model predictions to standardized emotion groups based on psychological emotion wheels, with the detailed mapping function described in the original paper~\cite{lian2025affectgpt}.
    \item For the MSA and OV-MER task~\cite{lian2025affectgpt}, we employ the Weighted Average F1-score (WAF) from MER-UniBench, which is well suited for multi-label classification scenarios.
    \item For the ERI task, evaluating free-form explanations requires semantic-level assessment. We adopt the Clue/Label Overlap metric from Emotion-LLaMA~\cite{cheng2024emotion}, which employs GPT-3.5-Turbo as an automatic judge to evaluate generated text in terms of multimodal cue completeness and emotion inference accuracy. Specifically, Clue Overlap measures the similarity between reasoning clues and ground truth, while Label Overlap assesses emotion recognition accuracy.
    \item For the MIR task, adhering to the official protocols of MIntRec~\cite{MIntRec} and MIntRec2.0~\cite{zhang2024mintrec}, we report accuracy (Acc), WAF, and weighted precision (WP).
    \item For the ERG task, we conduct a multifaceted evaluation. To measure whether the model's response is grounded in an accurate understanding of the user's emotion, we report both the fine-grained Acc from AvaMERG~\cite{zhang2025towards} and the coarse-grained Hit Rate from E3RG~\cite{10.1145/3746027.3762029}. To quantify the lexical diversity of the generated responses, we use Dist-n~\cite{li2015diversity}.
\end{enumerate}

\subsection{Human Blind Evaluation on the ERG task}
To ensure the reliability of automated evaluation metrics, we conducted a blind review by human experts for the empathetic generation task. Specifically, we randomly sample 200 dialogues (including the complete context of the conversation), and 10 human experts conduct blind evaluations using a 1 to 5 Likert scale on three metrics. As shown in \cref{tab: human-eval}, Nano\mbox{-}EmoX outperforms the baseline with an average Fleiss' Kappa of $\approx$ 0.697, achieving the best performance across all three dimensions and thus validating the reliability of automated metrics.
\begin{table}[h]
\centering
\footnotesize  
\caption{Human experts blind evaluation on the ERG task.}
\label{tab: human-eval}

\scriptsize 
\setlength{\tabcolsep}{6.5pt}
\begin{tabular}{c c c c c}
\toprule  
Models & Empathy $\uparrow$ & Insight $\uparrow$ & Safety $\uparrow$ & Avg.\\  
\midrule  
Qwen2.5-Omni-7B &3.98 & 4.03 & 4.59 &4.20\\
Ola-Omni-7B & 4.18 & 4.29 & 4.67 &4.38\\
\midrule
\rowcolor{gray!30} \multicolumn{5}{c}{Small-scale Multimodal Models} \\
MobileVLM V2-1.7B & 2.25 & 2.84 & 3.73 &2.94\\
AffectGPT (s) & 4.34 & 4.16 & 4.79 &4.43\\
\midrule
  \textbf{Our Nano\mbox{-}EmoX} & \textbf{4.75} & \textbf{4.42} & \textbf{4.87} &\textbf{4.68}\\  
\bottomrule  
\end{tabular}
\setlength{\tabcolsep}{6pt}
\end{table}



\subsection{Additional ablation study}
\paragraph{Ablation study on the fusion encoder.}
We investigated the impact of feature source depth by varying the number and position of the extracted encoder layers for fusion. As presented in \cref{tab:encoder_layer_ablation}, the results reveal that a three-layer configuration, sourcing from two intermediate layers (12, 16) and one deep layer (22), achieves the optimal performance. We observe that incorporating shallower features (e.g., from layer 8) provides limited benefits, likely due to their lack of semantic richness. Conversely, adding a fourth layer yields diminishing returns and fails to justify the increased computational cost. Thus, our three-expert setup strikes an effective balance between representational power and efficiency.
\begin{table}[t]
    \centering
    \footnotesize
    \caption{Exploring the appropriate number of experts and the depth of the extraction layer, extracting from too shallow a layer will lead to a decline in performance.}
    \setlength{\tabcolsep}{0.21pt} 
    \begin{tabular}{cccccc}
        \toprule
        \multirow{2}{*}{\makecell{Speech\\Extract Layers}} & 
        \multirow{2}{*}{\makecell{Visual\\Extract Layers}} & 
        \multirow{2}{*}{Expert} & 
        MSA\&MER\&OV-MER & ERI & ERG \\
        \cmidrule(lr){4-6} 
        & & &Avg. &Avg. &Hit Rate \\ 
        \midrule
        8 / 18          & 8 / 16          & 2 & 71.98 & 6.02 & 88.26 \\
        16 / 18           & 12 / 16           & 2 & 72.42 & 6.08 & 88.89 \\
        8 / 18 / 22      & 8 / 16 / 22      & 3 & 73.17 & 6.40 & 89.55 \\
        16 / 18 / 22     & 12 / 16 / 22     & 3 & \textbf{74.01} & \textbf{6.80} & \textbf{91.13} \\
        8 /16 / 18 / 22  & 8 / 12 / 16 / 22 & 4 & 71.09 & 5.70 & 91.12 \\
        \bottomrule
    \end{tabular}
    \label{tab:encoder_layer_ablation}
\end{table}
\paragraph{Ablation study on the vision token numbers.}
\cref{tab:token-visual} confirms that 32 tokens are sufficient for perception tasks. While increasing tokens benefits reasoning tasks, we selected 32 to achieve trade-off between efficiency and performance.
\begin{table}[h]
\centering
\footnotesize
\caption{The result of different visual token settings.}
\label{tab:token-visual}

\scriptsize 
\setlength{\tabcolsep}{4.0pt}
\begin{tabular}{c c c c c}
\toprule  
\multirow{2}{*}{Visual Tokens} & MSA\& MER\& OV-MER &MIR & ERI & ERG \\
\cmidrule{2-5}  
&Avg. $\uparrow$ & Avg. $\uparrow$ & Avg. $\uparrow$ & Hit Rate $\uparrow$ \\
\midrule  
32 & 74.01 & 52.72 & 6.80 & 91.13 \\
64 & 73.96 &55.48 & 6.83 & 91.08 \\
128 & 74.28 &60.53 & 6.95 & 92.87 \\
\bottomrule  
\end{tabular}
\setlength{\tabcolsep}{6pt}
\end{table}

\paragraph{Ablation study on task proportioning.}
We analyzed the task composition in Phase 3 of the P2E framework to identify the optimal training ratio for downstream tasks. As detailed in \cref{tab:task_ratio}, we identified a balanced configuration (MER:OV-MER:MIR:ERI:ERG = 18\%:28\%:5\%:31\%:18\%) that prioritizes foundational emotion perception and empathetic recognition. This comes at the acceptable cost of a minor performance dip in the MIR task. We posit that this is a favorable trade-off, as robust perceptual capabilities are a prerequisite for generating genuinely empathetic responses. This choice directly supports our overarching goal of bridging the cognitive gap from perception to empathy.
\begin{table*}[t]
    \centering
    \footnotesize
    \setlength{\tabcolsep}{3.5pt}
    \caption{Results of the ablation study on task composition in phase 3 of P2E. This table investigates the model's sensitivity to different proportions of training tasks.}
    \begin{tabular}{cccccc}
        \toprule
        \multirow{2}{*}{\makecell{P2E Phase3 Task Ratio\\(MER: OV-MER: MIR: ERI: ERG)}} & MER-UniBench & MIntRec & MIntRec 2.0 & EMER & AvaMERG \\
        \cmidrule(lr){2-6}
        & Avg. & WAF & WAF & Avg. & Hit Rate \\
        \midrule
        0\% : 20\% : 20\% : 25\% : 35\% & 71.43 \textcolor{Maroon}{(-2.58)} & 61.29 \textcolor{OliveGreen}{(+3.12)} & 49.8 \textcolor{OliveGreen}{(+2.53)} & 6.65 \textcolor{Maroon}{(-0.15)} & 43.15 \textcolor{Maroon}{(-44.03)} \\
        10\% : 30\% : 15\% : 35\% : 10\% & 72.79 \textcolor{Maroon}{(-1.22)} & 62.23 \textcolor{OliveGreen}{(+4.06)} & 51.04 \textcolor{OliveGreen}{(+3.77)} & 6.64 \textcolor{Maroon}{(-0.16)} & 58.88 \textcolor{Maroon}{(-28.3)} \\
        18\% : 20\% : 20\% : 25\% : 18\% & 72.60 \textcolor{Maroon}{(-1.41)} & 63.41\textcolor{OliveGreen}{(+5.24)} & 49.09 \textcolor{OliveGreen}{(+1.82)} & 6.60 \textcolor{Maroon}{(-0.20)} & 91.30 \textcolor{OliveGreen}{(+0.17)} \\
        18\% : 28\% : 5\% : 31\% : 18\% & 74.01 & 58.17 & 47.27 & 6.80 & 91.13 \\
        25\% : 17\% : 10\% : 22\% : 25\% & 72.18 \textcolor{Maroon}{(-1.83)} & 42.19 \textcolor{Maroon}{(-15.98)} & 52.09 \textcolor{OliveGreen}{(+4.82)} & 6.83 \textcolor{OliveGreen}{(+0.03)} & 87.18 \textcolor{Maroon}{(-3.95)} \\
        \bottomrule
    \end{tabular}
    \label{tab:task_ratio}
\end{table*}
\section{More visualization results}
\label{sup:more_visual_res}
\begin{figure*}[t] 
    \centering
    \includegraphics[width=1.0\textwidth]{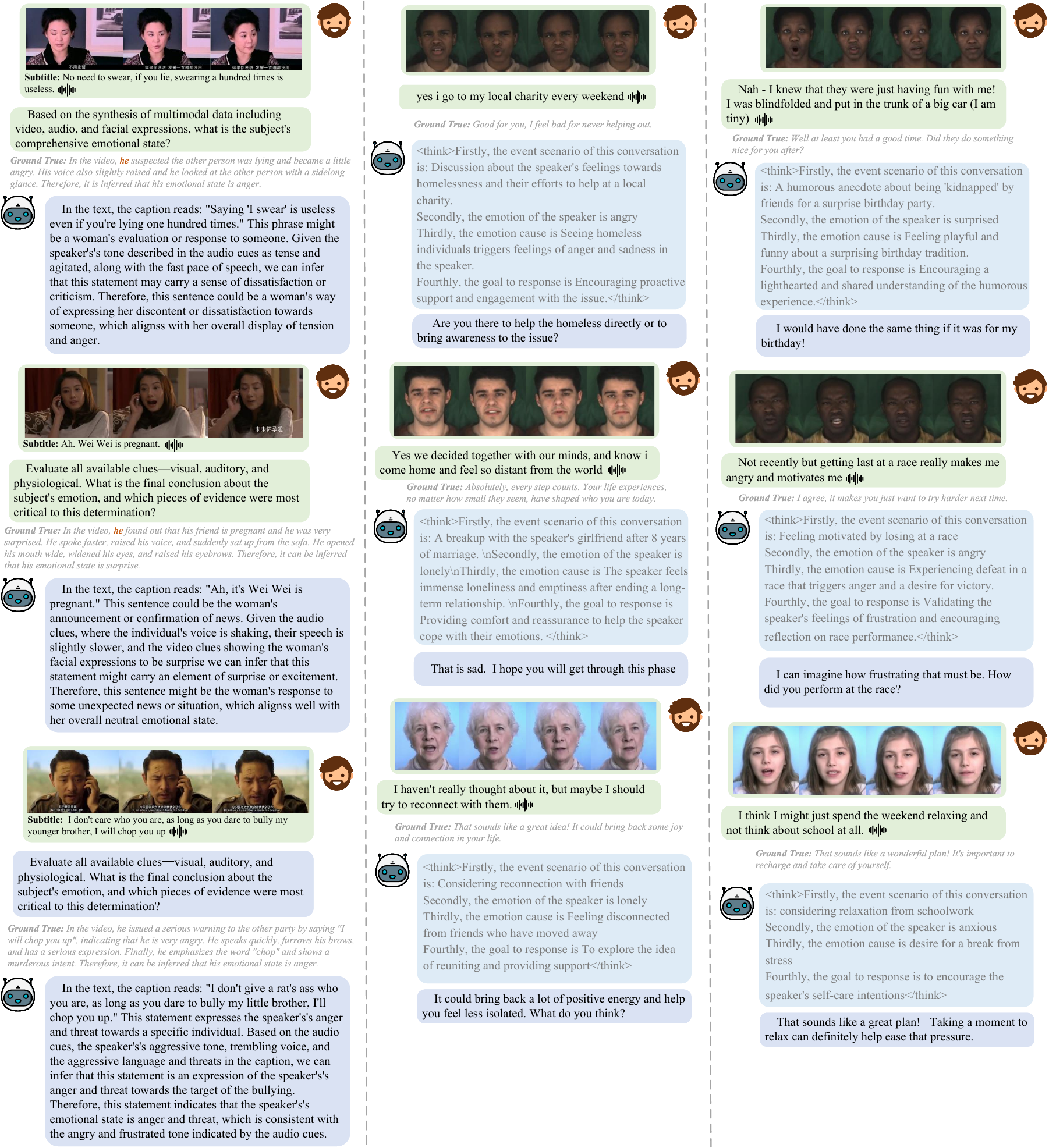}
    \caption{More visualization results in ERI and ERG task.}
    \label{fig:qualitative_analysis_additional}
\end{figure*}

We provide additional qualitative results to illustrate the interpretability and empathetic quality of Nano-EmoX's responses. In the \cref{fig:qualitative_analysis_additional}, our visualizations first demonstrate that the model can synthesize cues from visual, acoustic, and textual modalities to provide comprehensive causal explanations for an emotion. Furthermore, the model employs a multi-step reasoning process to progressively build an emotional context, which enables it to craft genuinely empathetic replies.
Taken together, these findings highlight Nano\mbox{-}EmoX's robust capabilities in both emotional understanding and empathetic interaction.
{
    \small
    \bibliographystyle{ieeenat_fullname}
    \bibliography{main}
}
\end{document}